\documentclass[sigconf]{acmart}

\usepackage{multirow}
\usepackage{subcaption}

\usepackage{graphicx}
\usepackage{epstopdf} 
\AtBeginDocument{%
  }

\begin{document}

\title{MGHFT: Multi-Granularity Hierarchical Fusion Transformer for Cross-Modal Sticker Emotion Recognition}

\author{Jian Chen}
\email{chenj589@mail2.sysu.edu.cn}
\authornotemark[1]
\affiliation{%
  \institution{Shenzhen MSU-BIT University}
  \city{Shenzhen}
  \state{Guangdong}
  \country{China}
}

\author{Yuxuan Hu}
\authornote{Both authors contributed equally to this research.}
\email{huyx55@mail2.sysu.edu.cn}
\affiliation{%
  \institution{Shenzhen MSU-BIT University}
  \city{Shenzhen}
  \state{Guangdong}
  \country{China}
}

\author{Haifeng Lu}
\email{luhf18@lzu.edu.cn}
\affiliation{%
  \institution{Shenzhen MSU-BIT University}
  \city{Shenzhen}
  \state{Guangdong}
  \country{China}
}
\author{Wei Wang}
\authornote{Corresponding authors.}
\email{ehomewang@ieee.org}
\affiliation{%
  \institution{Shenzhen MSU-BIT University}
  \city{Shenzhen}
  \state{Guangdong}
  \country{China}
}
\author{Min Yang}
\email{min.yang@siat.ac.cn}
\affiliation{%
  \institution{Shenzhen Institute of Advanced Technology, Chinese Academy of Sciences}
  \city{Shenzhen}
  \state{Guangdong}
  \country{China}
}

\author{Chengming Li}
\orcid{0000-0002-4592-3875}
\email{licm@smbu.edu.cn}
\authornotemark[2]
\affiliation{%
  \institution{Shenzhen MSU-BIT University}
  \city{Shenzhen}
  \state{Guangdong}
  \country{China}
}

\author{Xiping Hu}
\orcid{0000-0002-4952-699X}
\email{huxp@smbu.edu.cn}
\authornotemark[2]
\affiliation{%
  \institution{Shenzhen MSU-BIT University}
  \city{Shenzhen}
  \state{Guangdong}
  \country{China}
}

\begin{abstract}

Although pre-trained visual models with text have demonstrated strong capabilities in visual feature extraction, sticker emotion understanding remains challenging due to its reliance on multi-view information, such as background knowledge and stylistic cues. To address this, we propose a novel \textbf{m}ulti-\textbf{g}ranularity \textbf{h}ierarchical \textbf{f}usion \textbf{t}ransformer (\textbf{MGHFT}), with a multi-view sticker interpreter based on Multimodal Large Language Models. Specifically, inspired by the human ability to interpret sticker emotions from multiple views, we first use Multimodal Large Language Models to interpret stickers by providing rich textual context via multi-view descriptions. Then, we design a hierarchical fusion strategy to fuse the textual context into visual understanding, which builds upon a pyramid visual transformer to extract both global and local sticker features at multiple stages. Through contrastive learning and attention mechanisms, textual features are injected at different stages of the visual backbone, enhancing the fusion of global- and local-granularity visual semantics with textual guidance. Finally, we introduce a text-guided fusion attention mechanism to effectively integrate the overall multimodal features, enhancing semantic understanding. Extensive experiments on 2 public sticker emotion datasets demonstrate that MGHFT significantly outperforms existing sticker emotion recognition approaches, achieving higher accuracy and more fine-grained emotion recognition. Compared to the best pre-trained visual models, our MGHFT also obtains an obvious improvement, 5.4\% on F1 and 4.0\% on accuracy. The code is released at https://github.com/cccccj-03/MGHFT\_ACMMM2025.

\end{abstract}



\maketitle
\section{Introduction}
Stickers, as a popular and expressive form of online communication, serve as an important medium for users to convey emotions in online chatting~\cite{zhao2021affective}. Compared to plain text, stickers encapsulate a rich combination of visual elements and accompanying textual cues, enabling more vivid and nuanced emotional expression~\cite{tang2019emoticon,herring2017nice}. With the increasing integration of stickers into social media and instant messaging platforms, Sticker Emotion Recognition (SER) has emerged as a promising research direction, drawing growing interest from the academic community. SER aims to automatically identify and classify the emotional content conveyed through stickers by leveraging both visual and textual modalities~\cite{liu2022ser30k}. This research enhances natural and empathetic human-computer interaction while advancing affective computing with novel insights and practical implications for emotion-aware AI systems.

\begin{figure}[t]
    \centering
    \includegraphics[width=0.8\linewidth]{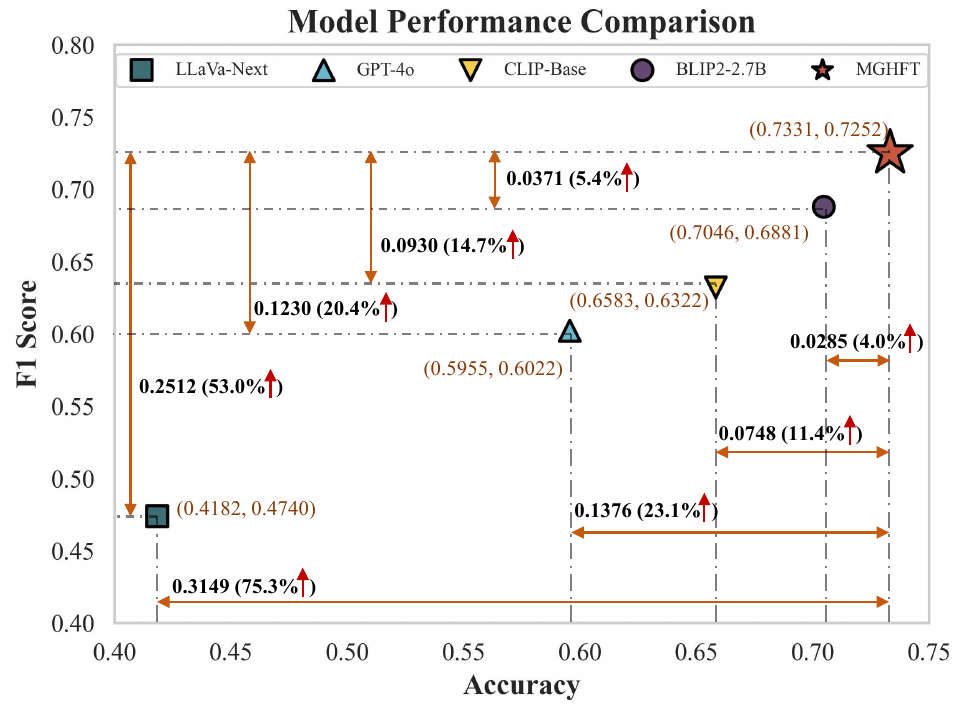}
    \vspace{-5mm}
    \caption{\textbf{Comparison with pre-trained models on SER30K dataset.} Multimodal Large Language Models such as LLaVA and GPT-4o can understand the content of stickers but cannot effectively recognize the emotion in them. Pre-trained visual models such as CLIP and BLIP can extract visual features for classification, but the performance is not good enough. Compared to these methods, our proposed MGHFT demonstrates an obvious improvement in both accuracy and F1 score.}
    \label{fig:vlmres}
    \vspace{-15pt}
\end{figure}


Compared to conventional image-based emotion recognition tasks~\cite{you2016building,yang2023emoset}, SER introduces a set of unique and more complex challenges. One of the most prominent difficulties lies in the implicit nature of emotional cues commonly found on stickers, such as culture background and style, which are often context-dependent and subtle~\cite{zheng2025multi,wang2024they}. Additionally, the vast diversity in sticker styles, themes, and visual representations further complicates the task of accurately identifying emotional content~\cite{chen2024tgca}. Unlike standard facial expression recognition or sentiment analysis from images, interpreting emotions in stickers often requires a holistic understanding that combines multiple views, including the intent behind the sticker~\cite{khattak2022efficient,ko2018brief}, its stylistic tone, character actions, and even cultural or conversational context. Even for humans, accurate emotion understanding of stickers frequently relies on background knowledge and contextual inference, underscoring the complexity of SER and the need for advanced multimodal modeling approaches.

Recent studies show that vision-language pre-trained models can effectively extract meaningful visual representations and have been widely applied to image understanding tasks~\cite{zhou2017places,he2016deep,radford2021learning,li2023blip}. On the one hand, multimodal foundation models such as LLaVA~\cite{liu2024llavanext} and GPT-4o~\cite{hurst2024gpt} excel in image-based generation tasks, leveraging their powerful visual-textual reasoning abilities and extensive background knowledge. On the other hand, vision-language models (VLMs) like CLIP~\cite{radford2021learning} and BLIP~\cite{li2023blip}, which are pre-trained through image-text alignment, have shown competitive performance in image classification tasks. Robust visual understanding provides a crucial foundation for the SER task, but it alone is not sufficient. More rich information about stickers from multi-views like style and details is also important. The results of the experiment presented in Figure~\ref{fig:vlmres} also prove this point. As illustrated in Figure~\ref{fig:vlmres}, experimental results on the SER30K dataset reveal a significant performance gap between these state-of-the-art VLMs and our proposed method. This discrepancy highlights an important insight.  Accurately perceiving and interpreting subtle, implicit emotional cues like intention remains a significant challenge. Therefore, achieving effective emotion recognition in stickers requires a more comprehensive integration of contextual knowledge, emotional reasoning, and multi-view understanding.

In this context, inspired by the way humans understand the emotion of stickers from multiple views, we propose a \textbf{M}ulti-\textbf{G}ranularity \textbf{H}ierarchical \textbf{F}usion \textbf{T}ransformer (\textbf{MGHFT}) to effectively integrate contextual text information into visual feature extraction. First, we introduce a multi-view sticker interpreter that leverages the powerful visual understanding capabilities of Multimodal Large Language Models (MLLMs) to generate rich textual descriptions from four views: intent, overall style, main character, and character details. These descriptions serve as rich semantic cues, aligning the model's emotional reasoning with human perception. Second, we develop a hierarchical fusion strategy that incorporates multi-view textual features into different stages of visual representation learning. Specifically, we adopt a Pyramid Vision Transformer (PVT) as the visual backbone and inject view-specific textual features at each layer. In every stage, a $CLS$ token captures global semantics, while local features are selectively attended based on attention weights. Through attention-based fusion and contrastive learning, the model performs fine-grained visual-textual alignment at both the global and local granularity. Finally, we introduce a Text-Guided Fusion Attention (TGFA) mechanism that aggregates and aligns multi-view textual semantics with visual representations across all stages, yielding robust emotion-aware features for SER. Extensive experiments on two large-scale sticker emotion datasets, SER30K and MET-MEME, show that our proposed MGHFT model consistently performs best, validating the effectiveness of MGHFT.

Our main contributions can be summarized as follows:

\begin{itemize}

\item We design a multi-view sticker interpreter that utilizes MLLMs to decompose stickers into multiple views, providing multi-view descriptions, thereby aligning human perceptual modalities to enrich the understanding of stickers. By introducing the knowledge of MLLMs, our model further improves the performance of emotion recognition.

\item We design a hierarchical fusion mechanism that injects multi-view textual semantics into different stages of visual feature extraction on the PVT backbone. By leveraging attention and contrastive learning at both global and local granularity, our approach enriches semantic representation and enhances emotion recognition performance.

\item We propose a novel text-guided multimodal fusion attention mechanism that leverages multi-view textual descriptions to guide the integration of textual and visual modalities, enhancing the model’s understanding of both the contextual knowledge and visual content of stickers.

\end{itemize}

\section{Related Work}

\begin{figure*}
    \centering
    \includegraphics[width=0.85\linewidth]{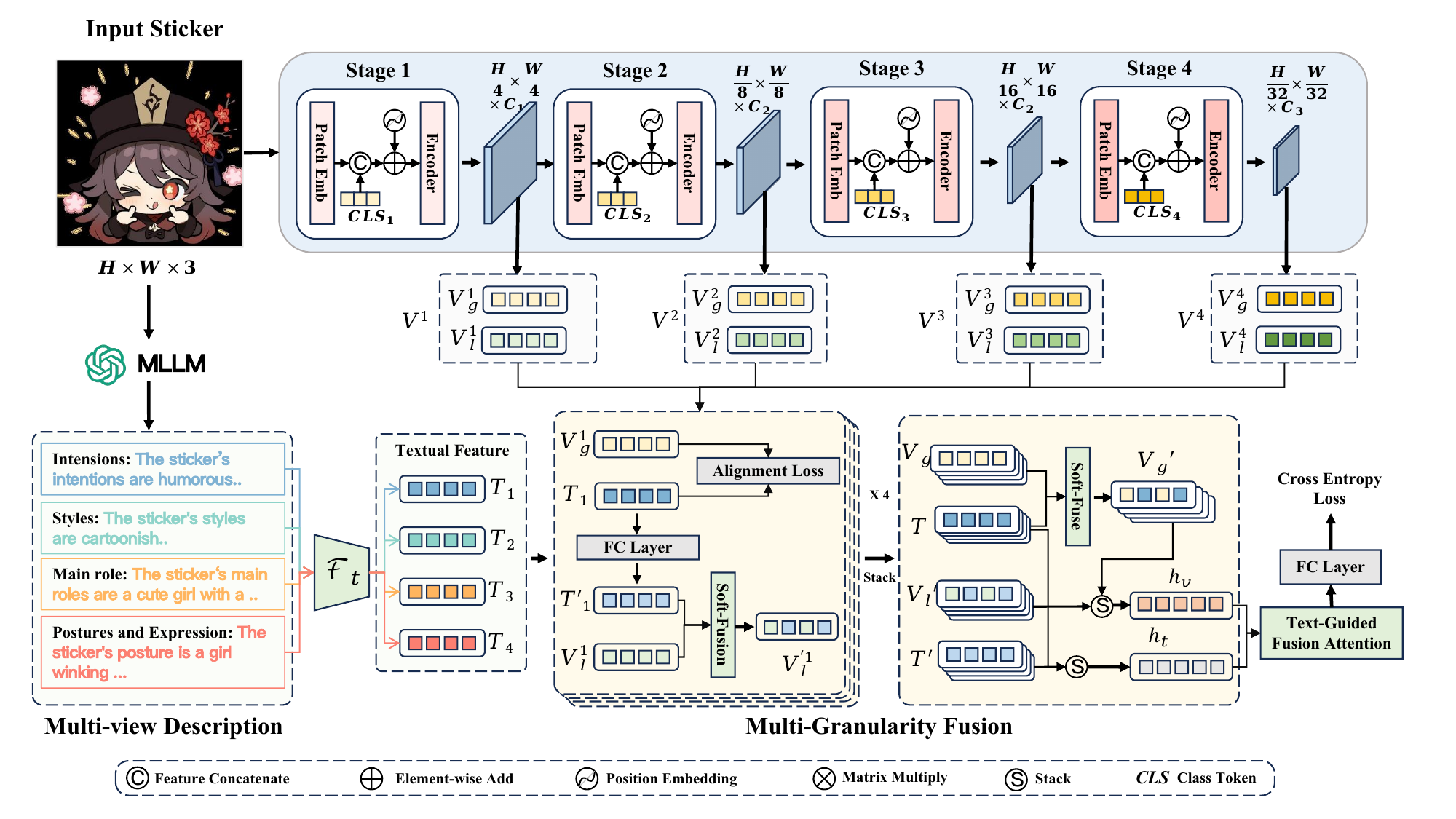}
    \vspace{-5mm}
    \caption{Framework of our proposed MGHFT. MGHFT adopts PVT as the backbone to extract the multi-granularity features of stickers $V_{g}$ and $V_{l}$. MLLM is used to generate multi-view descriptions of stickers. Alignment loss and the Soft-Fusion mechanism are proposed to fuse multi-granularity visual features with textual semantics in a hierarchical way.}
    \label{fig:framework}
\end{figure*}

\subsection{Sticker Emotion Recognition}

Accurate sticker emotion recognition remains a highly challenging task~\cite{derks2008role}. As one of the most popular forms of imagery in online conversations, stickers, particularly those with cartoon-style illustrations, enable users to convey emotions more effectively than plain text alone~\cite{wang2016more,10.1145/2957265.2961858,hu2024aptness}. However, due to the nature of stickers being sent as standalone messages, they are also more prone to emotional misinterpretation~\cite{cha2018complex}. 

Existing work has focused on annotated datasets and evaluation protocols for training sticker emotion recognition models. For instance, ~\citet{liu2022ser30k} constructs a large-scale dataset named SER30K specifically for this task, while ~\citet{xu2022met} focuses on collecting and organizing sticker data rich in metaphorical content. Other studies prioritize modeling relationships between multimodal features. For example, ~\citet{luo2024elemo} introduces overlapping patch embeddings to preserve local relationships within sticker images, enhancing the stability of feature representations. On the other hand, ~\citet{xu2024generating} extracts visual features from linguistic attributes and incorporates a text-conditioned generative adversarial network to better convey underlying language concepts. Furthermore, recognizing that stickers with similar themes often exhibit similar emotional characteristics, recent research begins to explore the role of thematic information in sticker emotion recognition. ~\citet{liu2022ser30k} extracts global and local thematic information and applies attention mechanisms for end-to-end sticker emotion recognition. ~\citet{chen2024tgca} further introduces a theme ID for each sticker and designs a theme-guided attention mechanism to enhance emotional recognition performance. Nonetheless, while these approaches achieve success in recognizing explicit emotional content in stickers, they still fall short in capturing deeper, implicit emotional cues such as metaphors and sarcasm, and that’s what we consider in our work.

\subsection{Cross-Modal Sticker Understanding}


In recent years, multimodal sticker understanding has attracted widespread attention~\cite{lin2024towards,kumari2024m3hop,cao2024modularized,jha2024memeguard,zhang2023skeletal}. Unlike general multimodal understanding tasks, sticker comprehension relies more heavily on contextual and external information~\cite{zheng2025multi}. 

Many multimodal fusion methods rely heavily on linguistic features extracted from OCR-recognized text. For example, references~\cite{duan2022browallia,zhuang2022yet} employ LSTM networks to encode textual information obtained from OCR. ~\citet{wang2024they}introduced both cross-modal and intra-modal attention mechanisms, along with a multimodal matching loss, to better capture the interactions between text and image for enhanced multimodal sticker understanding. ~\citet{zheng2025multi} focused on finer-grained features and proposed an object-level multimodal interaction framework. Considering that stickers frequently appear in dialogue scenarios, some works have integrated sticker understanding into conversational contexts. For instance, certain studies incorporate causal knowledge from text to help models recognize the emotional states expressed in stickers, thereby improving understanding performance~\cite{chen2024deconfounded}. Other approaches leverage large-scale knowledge learning and distillation to enrich the model’s feature representations to further enhance sticker comprehension~\cite{xia2024perceive}. Additionally, some research emphasizes dialogue intent recognition, using intent-driven alignment mechanisms to achieve deeper understanding of stickers within conversations~\cite{liang2024reply}. Distinct from these methods, this study focuses on leveraging multi-view emotional cues to guide multimodal fusion, aiming to enhance the model's ability to comprehend stickers more effectively.

\section{Methodology}
\noindent
\textbf{Problem Definition.} SER is formulated as an image classification task. The input is a sticker image of size $H \times W \times 3$, where $H$ and $W$ denote the height and width, respectively, and 3 represents the RGB channels. Our objective is to feed a sticker into MGHFT, which leverages the capabilities of MLLMs to interpret the sticker from multiple views and accurately predict its emotional category.

\noindent
\textbf{Overview.} Our proposed MGHFT method can be divided into three main parts. First, we adopt MLLM to obtain multi-view description information of stickers to align with how humans understand stickers. Then, we use PVT to extract the multi-granularity features of the stickers in stages, and we design a novel multi-granularity cross-modal fusion mechanism to integrate the descriptive information from different views into different stages for better representation. Finally, a topic-guided fusion attention is proposed to fuse the visual features with textual features. The whole framework of our proposed MGHFT can be seen in Figure~\ref{fig:framework}.

\subsection{Vision Backbone}

PVT is a widely used vision backbone for extracting multi-scale image features, and we adopt it to support our hierarchical cross-modal fusion approach. PVT contains four stages, which generate feature maps at various scales through an asymptotic reduction strategy applied at each stage's patch-embedding layer. It also leverages a self-attention mechanism to preserve a global receptive field. Consequently, at each stage $i$, PVT produces global features $V_{g}^{i}$, which are CLS tokens that capture aggregated global information, and local features $V_{l}^{i}$, which are key tokens selected based on attention weights \cite{liu2022ser30k}.


\subsection{MLLM-Based Multi-View Sticker Interpreter}



When encountering a new sticker, humans typically interpret it from multiple views. On the one hand, humans will consider the intended usage scenario~\cite{park2024memeintent,shifman2013memes} and the overall visual style~\cite{biederman1987recognition,shifman2013memes,zhang2024stickerconv} to form an initial understanding of its meaning. On the other hand, humans pay attention to the specific characters and their detailed features to assess whether the sticker suits their communicative needs~\cite{nguyen2024computational,sharma2023you,kress2020reading,liang2024reply}. Inspired by this human sticker comprehension process, we divide sticker understanding into four views. To help the model better understand the contextual information of stickers, we designed an MLLM-based Multi-View Sticker Interpreter. This module extracts multi-perspective information from stickers and incrementally injects it into the model through a multi-stage fusion process. Specifically, we define four essential views to comprehensively capture sticker semantics: intention $D_{I}$, overall style $D_{S}$, main roles $D_{MR}$, and fine-grained character details $D_{PE}$. These views enable a more holistic and detailed interpretation of sticker content. In recent years, MLLMs have demonstrated remarkable performance in the field of image understanding. These models can effectively activate their deep feature extraction capabilities through carefully designed prompts, enabling precise semantic comprehension of visual content. To enhance the semantic representation of sticker images, we employ the MLLM LLaVA-NeXT\footnote{https://huggingface.co/llava-hf/llava-v1.6-mistral-7b-hf}~\cite{liu2024llavanext}, leveraging its powerful cross-modal understanding ability to generate multi-view attribute descriptions of stickers.

\begin{equation}
    \{C_{I}, C_{S}, C_{MR}, C_{PE}\} = MLLM(\{D_{I}, D_{S}, D_{MR}, D_{PE}\})
    \label{eq:mllm}
\end{equation}


\begin{figure}
    \centering
    \includegraphics[width=0.7\linewidth]{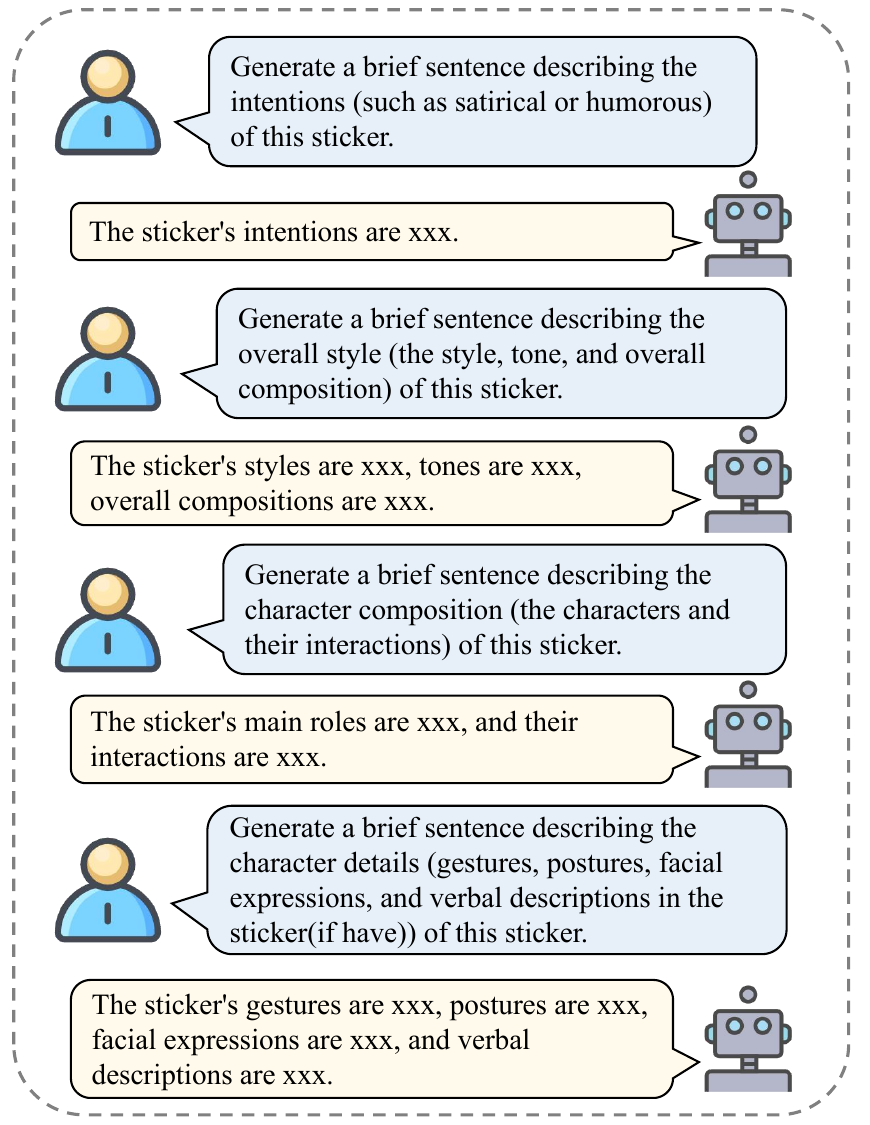}
        \vspace{-4mm}
    \caption{Multi-view sticker description generation.}
    \label{fig:prompt}
\end{figure}

As illustrated in Figure~\ref{fig:prompt}, we adopt a multi-round interaction strategy with an MLLM, leveraging carefully crafted prompts to direct its attention to specific aspects of each sticker view. This process facilitates the generation of fine-grained and informative descriptions across multiple views. Subsequently, we adopt a BERT-based text encoder as the textual backbone to convert these attribute descriptions into dense textual representations for downstream processing as Eq.~\ref{eq:bert}.

\begin{equation}
    T=\{T_{1}, T_{2}, T_{3}, T_{4}\} = BERT(\{C_{I}, C_{S}, C_{MR}, C_{PE}\})
    \label{eq:bert}
\end{equation}

\subsection{Multi-Granularity Cross-Modal Fusion}

After obtaining multi-view textual representations, our goal is to integrate this knowledge into the visual feature extraction process, guiding the model to focus on relevant visual cues. Specifically, we design a novel hierarchical fusion mechanism that injects multi-view knowledge into different stages of visual representation learning. Considering that textual descriptions may capture both global and local aspects of a sticker, we introduce a multi-granularity fusion strategy to ensure effective alignment and integration between textual features and visual representations.

On the one hand, local features $V_{l}$ correspond to key visual tokens, and directly aligning them with textual features may lead to the loss of important visual semantics. To address this, we adopt a fusion strategy at the local granularity level, integrating textual and local visual features to obtain richer representations. On the other hand, as global features encapsulate the overall semantics of a sticker, aligning and integrating them with multi-view textual features helps the model better understand emotional information and learn more expressive global representations.
 To bridge the semantic gap between global visual features $V_{g}$ and textual features $T$, we introduce an alignment loss. Furthermore, after completing all stages of feature extraction, we perform a final fusion between the overall global features and textual features to mitigate the potential performance impact caused by the sequential order of multi-view integration.


\subsubsection{\textbf{Local-Granularity.}}  For local features $V^{i}_{l}$, we adopt a Soft-Fusion attention mechanism to fuse it with textual features. Specifically, we first use linear layers to project the textual feature $T_{i}$ into $T'_{i}$, which has the same dimension of local-granularity vision features $V^{i}_{l}$. Then, we compute the attention scores between $V^{i}_{l}$ and $V'^{i}_{t}$ and apply softmax normalization to the scores. Then, it performs a weighted summation to enhance the representation of local features. The whole process can be illustrated in Eq.~\ref{eq:softfuse}.
\begin{equation}
    V'^{i}_{l} = V^{i}_{l} + \text{Softmax}(V^{i}_{l}\cdot {T'_{i}}^{T})\cdot T'_{i}
    \label{eq:softfuse}
\end{equation}

In this way, the proposed Soft-Fusion mechanism effectively leverages additional information from $T_{i}$ and utilizes the attention mechanism to enable $V^{i}_{l}$ to focus on the most relevant parts of $T_{i}$, significantly enriching the feature representation of $V^{i}_{l}$. Moreover, we introduce a residual connection to prevent gradient vanishing, ensuring that the original feature information is preserved and facilitating more robust learning. We adopt this mechanism in each stage to hierarchically fuse the different views of textual features into the local-granularity visual features. Then, we use MLPs to project the $V'^{i}_{l}$ and $T'_{i}$ into the output dimension for further fusion.

\subsubsection{\textbf{Global-Granularity.}}
For global features $V^{i}_{g}$, we employ an alignment loss to leverage textual representations as guidance for global visual feature extraction, encouraging the model to learn cross-modal features effectively. This loss consists of two components: Contrastive Loss $\mathcal{L}_{cl}$ and Multi-Level Cross-Entropy Loss (MLCE Loss) $\mathcal{L}_{mlce}$ \cite{yang2024accurate}. Together, they facilitate the alignment of visual and textual features, thereby enhancing the cross-modal representations effectively. We normalize the global visual and textual features to obtain the contrastive loss as Eq.~\ref{eq:loss_con}.
\begin{equation}
\begin{aligned}
    f_v &=  \frac{V^i_g}{\|V^i_g\|}, \quad f_t = \frac{T_i}{\|T_i\|},\quad S_{vt} =  \frac{f_v f_l^T}{\tau}, \\
    &\mathcal{L}_{cl} = \frac{1}{2} \left( \mathcal{L}_{ce}(S_{vt}, \mathbf{y}) + \mathcal{L}_{ce}(S_{vt}^T, \mathbf{y}) \right),
\end{aligned}
\label{eq:loss_con}
\end{equation}
where \(S_{vt}\) is the similarity between global visual and textual features, \(\tau\) is a temperature parameter, \(\mathcal{L}_{ce}\) denotes the Cross-Entropy Loss, and \(\mathbf{y}\) represents the matching indices. 

Furthermore, we introduce the MLCE Loss to improve the visual-text alignment robustness, which utilizes cosine similarity and Kullback-Leibler Divergence $D_{\text{KL}}$ to distill information between features as Eq.~\ref{eq:mlce}.
\begin{equation}
\begin{aligned}
    C_v &= 0.5 \left( 1 + f_v f_v^T \right), \quad C_t = 0.5 \left( 1 + f_l f_l^T \right), \\
    W_l &= \text{Softmax} \left( \frac{C_l}{\tau} \right), \quad W_h = \text{Softmax} \left( \frac{C_h}{\tau} \right), \\
    & \ \ \ \ \ \ \ \ \ \ \ \ \ \mathcal{L}_{mlce} = D_{\text{KL}}(W_l \| W_h),
\end{aligned}
\label{eq:mlce}
\end{equation}
where $C_l$ and $C_h$ are the normalized self-similarity matrices of
The final cross-modal alignment loss is given by Eq.~\ref{eq:loss_align}.
\begin{equation}
\mathcal{L}_{align} = \mathcal{L}_{cl} + \lambda \mathcal{L}_{mlce},
\label{eq:loss_align}
\end{equation}
where \(\lambda\) is set as 30 to control the contribution of MLCE loss to the total alignment loss \cite{yang2024accurate}. 

We then separately stack the global features $h_{g}$ and textual features $h_{t}$ and apply the Soft-Fusion Mechanism to enhance the global-granularity features, resulting in $h'_{g}$, as shown in Eq.~\ref{globalfuse}.
\begin{equation}
    h'_{g} = h'_{g} + \text{Softmax}(h'_{g}\cdot {h_{t}}^{T})\cdot h_{t}
    \label{globalfuse}
\end{equation}

Ultimately, for both textual and visual features, we obtain corresponding local-granularity and global-granularity representations, which are then separately stacked for the final cross-modal fusion.

\subsection{Text-Guided Fusion Attention}
\begin{figure}
    \centering
    \includegraphics[width=0.7\linewidth]{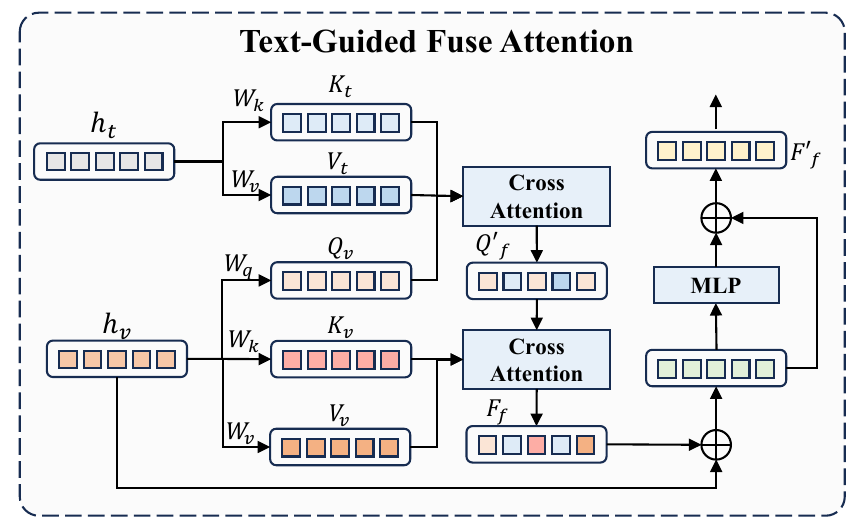}
    \vspace{-4mm}
    \caption{Framework of TGFA for cross-modal fusion.}
    \label{fig:tgfa}
\end{figure}

To better guide the integration of visual representations with textual features, we propose a Text-Guided Fusion Attention (TGFA) mechanism, as shown in Figure~\ref{fig:tgfa}.

Specifically, given visual and textual features, we first apply cross-attention, where $h_{v}$ is used to generate the query $Q_{v}$ while $h_{t}$ produces the key $K_{t}$ and value $V_{t}$. The output, denoted as $Q'_{f}$, integrates cross-modal features. Next, we use $Q'_{f}$ as the query for a second cross-attention step, with $h_{v}$ providing the key $K_{v}$ and value $V_{v}$. To further enhance cross-modal feature fusion, we incorporate an MLP and a residual connection. The complete process of TGFA is illustrated in Eq.~\ref{eqTGFA}.

\begin{equation}
    \begin{aligned}
        Q'_{f} &=  \text{Softmax}(\frac{Q_{v}K_{t}^{T}}{\sqrt{d_{head}}})V_{t}, \\
        F_{f} &= \text{Softmax}(\frac{Q'_{f}K_{v}^{T}}{\sqrt{d_{head}}})V_{v}, \\
        F'_{f} &= MLP(h_{v} + F_{f}) + h_{v} + F_{f}
    \end{aligned}
    \label{eqTGFA}
\end{equation}
where $Q_{j}=W_{q}h_{j}+b$,$K_{j}=W_{k}h_{j}+b$,$V_{j}=W_{v}h_{j}+b$ are linear transformations. The final output, $F'_{f}$, is then fed into an FC layer for emotion recognition. It is also worth noting that we employ multi-head attention, and Eq.~\ref{eqTGFA} illustrates the process for each individual attention head. We then use Cross Entropy Loss $\mathcal{L}_{CE}$ to optimize the model. The whole loss function of model training is shown as Eq.~\ref{eqloss}.
\begin{equation}
    \mathcal{L}_{total} = \mathcal{L}_{CE} + 0.5 *\mathcal{L}_{align},
    \label{eqloss}
\end{equation}
where the weight of alignment loss is set as 0.5 to enhance the classification task.

\section{Experiments}

\begin{table*}[htbp]
  \centering
  \caption{Performance in the SER30K dataset for SER. The result in bold is the best, while the underline is the second best.}
  \vspace{-3mm}
  \scalebox{0.8}{
    \begin{tabular}{c|l|ccccccc|c|c}
    \toprule
    \multirow{2}[4]{*}{Method} & \multicolumn{1}{c|}{\multirow{2}[4]{*}{Model}} & \multicolumn{7}{c|}{Precision on each emotion category} & \multicolumn{1}{c|}{\multirow{2}[4]{*}{Accuracy}} & \multirow{2}[4]{*}{F1} \\
\cmidrule{3-9}          & \multicolumn{1}{c|}{} & Anger & Disgust  & Fear  & Happiness & Neutral & Sadness & Surprise &       &  \\
    \midrule
    \midrule
    \multirow{8}[2]{*}{Image} 
          & VGG \cite{zhou2017places} & 37.40  & 0.00  & 41.02  & 73.55  & 60.76  & 51.52  & 35.14  & 62.57  & - \\
          & ResNet \cite{he2016deep} & 50.30  & 26.66  & 58.01  & 76.63  & 65.94  & 64.69  & 48.33  & 67.76  & - \\
          & ViT \cite{dosovitskiy2020image}  & 52.72  & 32.00  & 53.70  & 74.68  & 62.92  & 56.55  & 40.31  & 64.94  & - \\
          & WSCNet \cite{yang2018weakly} & 58.77  & 0.00  & \textbf{74.62}  & 79.49  & 63.50  & 65.96  & 49.53  & 68.58  & - \\
          & PDANet \cite{zhao2019pdanet} & 58.10  & 26.66  & 61.68  & \underline{79.60}  & 64.76  & 63.50  & 47.10  & 68.68  & - \\
          & LORA \cite{liu2022ser30k} & 54.71  & \underline{50.00}  & 64.15  & 78.04  & 67.03  & 66.25  & 44.68  & 69.22  & 68.93  \\
          & MAM \cite{zhang2024affective}  & 55.67  & 0.00  & \underline{71.88}  & 78.50  & 65.26  & 61.27  & \textbf{59.42}  & 68.97  & 67.60  \\
            & TGCA-PVT \cite{chen2024tgca} & 57.51  & \textbf{57.14}  & 68.42  & 76.92  & 68.96  & 65.48  & 52.40  & 70.23  & 69.85  \\
    \midrule
    \multirow{6}[2]{*}{Image+Text} 
    & WSCNet \cite{yang2018weakly} & 56.64  & 36.84  & 60.00  & 77.85  & 66.72  & 67.04  & 49.18  & 69.45  & - \\
    & PDANet \cite{zhao2019pdanet} & 60.09  & 19.23  & 59.29  & 80.57  & 65.02  & 61.08  & 48.98  & 68.93  & - \\   
          & LORA \cite{liu2022ser30k} & 60.18  & \underline{50.00}  & 65.09  & 76.83  & 68.78  & \underline{67.70}  & \underline{55.14}  & 70.73  & 69.91  \\
          & MAM+Bert \cite{zhang2024affective} & 60.82  & 1.00  & 65.28  & 79.09  & 65.98  & 61.20  & 53.51  & 69.75  & 68.58  \\
          & TGCA-PVT \cite{chen2024tgca} & \underline{65.67}  & 35.73  & 66.09  & 79.57  & \underline{69.39}  & 63.62  & 53.04  & \underline{71.63}  & \underline{70.93}  \\
     & \textbf{MGHFT} (ours) & \textbf{68.58}  & 22.22  & 62.71  & \textbf{79.79}  & \textbf{70.82}  & \textbf{71.80}  & 52.47  & \textbf{73.31 } & \textbf{72.52 } \\
    \bottomrule
    \end{tabular}%
    }
  \label{tab:serres}%
\end{table*}%
\subsection{Datasets}
\noindent
\textbf{SER30K.} SER30k \cite{liu2022ser30k} contains 30,739 stickers with 7 emotional categories. The stickers are crawled from the sticker image website\footnote{https://getstickerpack.com}, while the labels are annotated by three annotators. Stickers that are non-English or dynamic format are removed during collection. For data splitting, we adopt 7:1:2 for training, validation, and testing.

\noindent
\textbf{MET-MEME.} MET-MEME \cite{xu2022met} contains both English version and Chinese version. Considering the consistency between MET-MEME and SER30K and common MLLMs, which are pre-trained in English, we adopted the English version here for experiments. The English
version is sourced from MEMOTION and Google search, including 4,000 stickers with 7 emotional categories. For data splitting, we adopt 6:2:2 for training, validation, and testing.

\subsection{Experimental Setting}
\noindent
\textbf{Implementation Details.} 
We adopt the Pytorch framework to conduct all experiments on 2 NVIDIA A800 80GB GPUs. We adopt the pre-trained PVT-small model \cite{wang2021pyramid} as the visual backbone. The max length of the sequence features obtained by the pre-trained Bert model \cite{abs-1810-04805} is set as 512. We adopt AdamW to optimize the model with a learning rate of 1e-3. The epoch is set as 50, while the batch size is set as 16. It should be noted that, adopting the same setting of LORA \cite{liu2022ser30k}, the parameters of the Bert model are frozen, while the parameters of the PVT model are trainable.

\noindent
\textbf{Evaluation metrics.} Since SER is a multi-class classification task, we adopt precision, recall, accuracy, and F1 score as evaluation metrics. We also provide the precision of each category on SER30K for better comparison. Considering that SER30K is a large-scale dataset, we conduct most experiments on this dataset and use MET-MEME as a complement to robustness validation.

\subsection{Comparison Results}
On the SER30K dataset, we conduct two sets of comparative experiments: (1) compared with SER models or image classification models and (2) compared with pre-trained visual models. For (1), we adopt baselines include: VGG \cite{zhou2017places} and ResNet \cite{he2016deep}, ViT \cite{dosovitskiy2020image}, WSCNET \cite{yang2018weakly}, PDANet \cite{zhao2019pdanet}, LORA \cite{liu2022ser30k}, MAM \cite{zhang2024affective}, and TGCA-PVT \cite{chen2024tgca}. We followed the baseline result of Ref.~\cite{liu2022ser30k}. For (2), we adopt CLIP-Base \cite{radford2021learning}, BLIP2-2.7B \cite{li2023blip}, LLaVA-NeXT-Mistral-7B \cite{liu2024llavanext} and GPT-4o \cite{hurst2024gpt}. For CLIP-Base and BLIP2-2.7B, only a classifier head is retrained for classification. 

Table~\ref{tab:serres} shows the emotion recognition results of (1) on the SER30K dataset with SER models. According to the comparison results shown in the table, our proposed MGHFT demonstrates a significant performance advantage on the SER30k dataset, achieving the highest accuracy of 73.31\% and an F1 score of 72.52\%. Compared to the best-performing baseline method, TGCA-PVT, which achieves 71.63\% accuracy and 70.93\% F1 score, MGHFT improves the accuracy by 2.3\% and the F1 score by 2.2\%. Compared to the image emotion recognition method MAM, our proposed MGHFT shows a much more significant improvement. These consistent improvements across both metrics highlight the effectiveness of our method in more comprehensively capturing the emotion-related features of stickers.

In particular, the results of (2) are shown in Figure~\ref{fig:vlmres}. The much lower performance results than other methods suggest that LLaVA and GPT-4o are still struggling with the task of directly understanding sticker emotions. In addition, CLIP and BLIP2, with their strong image feature extraction capability, achieved a relatively good performance after training on the classification head, but there is still a gap in comparison to the models specifically used for sticker emotion understanding. Compared with these pre-trained visual models, our proposed MGHFT model demonstrates significant advantages in emotion recognition results, achieving an improvement of 5.4\% in F1 score and 4.0\% in accuracy over the BLIP2, which is the best-performing pre-trained model.

\begin{table}[htbp]
  \centering
  \caption{Performance on \textbf{MET-MEME} dataset.}
  \vspace{-4.5mm}
  \scalebox{0.8}{
    \begin{tabular}{lccc}
    \toprule
    Method & Accuracy   & Precision   & Recall \\
    \midrule
    \midrule
    VGG15~\cite{simonyan2014very} & 20.57  & 20.84  & 24.22  \\
    DenseNet-161~\cite{huang2017densely} & 21.88  & 21.71  & 25.65  \\
    ResNet-50~\cite{he2016deep} & 21.74  & 18.63  & 21.35  \\
    Multi-BERT\_EfficientNet~\cite{tan2019efficientnet} & 28.52  & 24.52  & 29.04  \\
    Multi-BERT\_ViT~\cite{dosovitskiy2020image} & 24.43  & 23.41  & 23.96  \\
    Multi-BERT\_PiT~\cite{heo2021rethinking} & 25.00  & 27.82  & 28.12  \\
    MET\_add~\cite{xu2022met} & 24.65  & 24.52  & 25.26  \\
    MET\_cat~\cite{xu2022met} & 27.68  & 28.41  & 29.82  \\
    M3F\_add~\cite{wang2024they} & 30.47  & 33.45  & 30.34  \\
    M3F\_cat~\cite{wang2024they} & 29.82  & 34.18  & 30.73  \\
    MGMCF~\cite{zheng2025multi} & \underline{34.36}  & \textbf{37.77 } & \underline{34.38}  \\
    MGHFT (ours) & \textbf{35.13} & \underline{34.75}  & \textbf{35.12} \\
    \bottomrule
    \end{tabular}%
    }
  \label{tab:met}%
\end{table}%

To further evaluate the robustness of our proposed MGHFT, we conduct experiments on the MET-MEME dataset, as shown in Table~\ref{tab:met}. We adopt the baselines and experimental results from Ref.~\cite{zheng2025multi}. Compared to the SER30k dataset, emotion recognition on MET-MEME is more challenging due to the significantly smaller number of training samples and greater visual-textual variation. Despite the increased difficulty, MGHFT still achieves the best performance across all evaluation metrics, with an accuracy of 35.13\%, a precision of 34.75\%, and a recall of 35.12\%. Compared with the strongest baseline method, MGMCF, which achieves a precision of 34.36\%, a precision of 37. 77\% and a recall of 34.88\%, our method achieves a 0.77 percentage point gain in accuracy while maintaining a comparable performance in recall and more balanced precision. These results demonstrate that MGHFT not only achieves state-of-the-art performance on large-scale datasets like SER30K but also maintains strong generalization and robustness in low-resource, more challenging scenarios such as MET-MEME.

\subsection{Ablation Study}
To verify the effectiveness of each individual component in our proposed MGHFT model, we conduct comprehensive ablation studies. Specifically, we perform two types of experiments: retaining each module individually to assess its standalone contribution and removing each module separately to evaluate its impact on overall performance. \textbf{CL} means the contrastive learning between $V_{g}$ and $T_{i}$ used in our method, \textbf{TGFA} denotes our proposed Text-Guided Fusion Attention module, while \textbf{GF} and \textbf{LF} represent the soft fusion mechanisms for global-granularity and local-granularity feature fusion, respectively. Results are shown in Table~\ref{tab:ab1}.

\begin{table}[htbp]
  \centering
  \caption{Ablation results for each module on the SER30K.}
  \vspace{-4.5mm}
    \setlength{\tabcolsep}{8pt} 
    \scalebox{0.85}{
    \begin{tabular}{cccccc}
    \toprule
    CL & GF    & LF    & TGFA  & Accuracy   & F1 \\
    \midrule
    \midrule
         &       &       &       & 70.07  & 69.89  \\
    $\surd$     &       &       &       & 71.65  & 70.98  \\
          & $\surd$     &       &       & 71.63  & 70.76  \\
          &       & $\surd$     &       & 71.67  & 70.80  \\
          &       &       & $\surd$     & 71.13  & 70.78  \\
    $\surd$     & $\surd$     & $\surd$     &       & 72.61  & 71.54  \\
    $\surd$     & $\surd$     &       & $\surd$     & 73.03  & 72.19  \\
    $\surd$     &       & $\surd$     & $\surd$     & 72.98  & 72.39  \\
    & $\surd$     & $\surd$     & $\surd$     & 72.23  & 71.59  \\
    $\surd$     & $\surd$     & $\surd$     & $\surd$     & \textbf{73.31 } & \textbf{72.52 } \\
    \bottomrule
    \end{tabular}%
    }
  \label{tab:ab1}%
\end{table}%

As shown in Table~\ref{tab:ab1}, none of the ablated variants outperform the full model, which underscores the complementary roles and necessity of each component in achieving optimal performance. For the model with a single module, we can observe that the model with only the TGFA module exhibited the most significant performance drop compared to the full model. Compared to the backbone without any module, the model with CL has the highest F1 score  (from 69.89\% to 70.98\%). This result is also consistent with the results of removing each module from the full model: the model without CL has the most significant performance degradation compared to the full model than with the other modules removed.

\begin{figure}[h]
    \centering
    \includegraphics[width=0.85\linewidth]{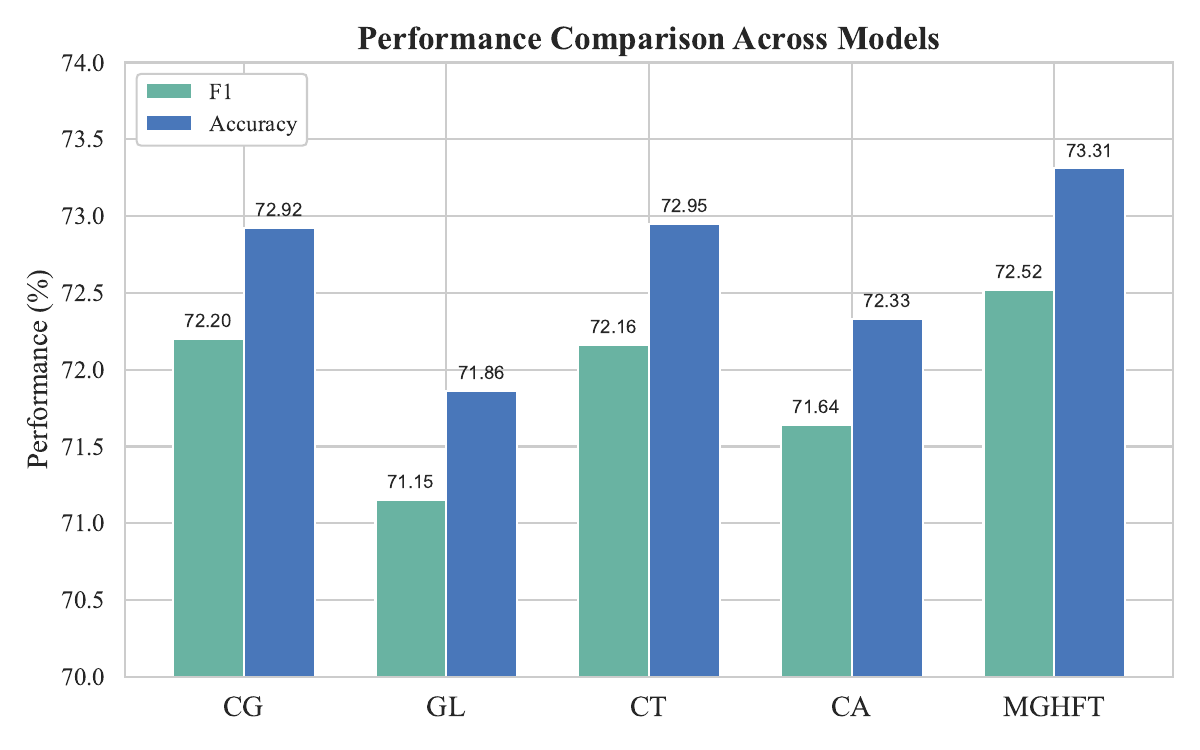}
    \vspace{-4.5mm}
    \caption{Performance of each model variants.}
    \label{fig:ab3}
\end{figure}

Furthermore, we evaluate combinations of two components to explore their joint effect. We also adopt the commonly used cross-attention module to replace our proposed Soft-Fusion mechanism (CA). \textbf{CG} refers to using \textbf{CL} and \textbf{GF}, \textbf{GL} refers to using \textbf{GF} and \textbf{LF}, and \textbf{CT} refers to using \textbf{CL} and \textbf{TGFA}. The performance of each variant is shown in Figure~\ref{fig:ab3}.

The results indicate that combining any two of the proposed modules consistently enhances sticker emotion classification performance. Notably, the combinations of CL with GF and CL with TGFA yielded the most significant improvements. Specifically, CL helps align multi-view textual features with global visual representations, effectively narrowing the semantic gap between modalities and enhancing the performance of both GF and TGFA modules. Interestingly, the combination of GF and LF, without contrastive learning, resulted in a more substantial improvement in the F1 score than accuracy. This suggests that multi-granularity feature fusion improves the model’s ability to recognize diverse emotions, leading to more balanced classification outcomes.  The results of CA also show that our proposed Soft-Fusion not only achieves better fusion performance but also avoids introducing extra parameters, thereby improving computational efficiency.

\subsection{Analysis of Text Description}
Moreover, we conduct experiments to further evaluate the effectiveness of the proposed multi-view textual descriptions in sticker emotion understanding. Given that our method leverages contextual descriptions encompassing multiple views, it is essential to investigate how the integration of various views at different stages influences performance. Additionally, to assess whether the diverse views contribute richer contextual information, we replace all views with only the character details ($T_{4}$) and evaluate the impact. Furthermore, we concatenate the descriptions from all views to form a unified multi-view representation $T$, integrating it at each stage of the process. The experimental results, as presented in Table~\ref{tab:abseq}, demonstrate the effectiveness and importance of incorporating multi-view textual information.

\begin{table}[htbp]
  \centering
  \caption{Effect of multi-view text description.}
  \vspace{-4.5mm}
  \scalebox{0.85}{
    \begin{tabular}{cccccc}
    \toprule
    Stage 1 & Stage 2 & Stage 3 & Stage 4 & Accuracy   & F1 \\
    \midrule
    \midrule
    $T_{1}$     & $T_{2}$      & $T_{3 }$     & $T_{4}$      & \textbf{73.31 } & \textbf{72.52 } \\
    $T_{1}$      & $T_{2}$      & $T_{4}$      & $T_{3}$      & 73.02  & 72.39  \\
    $T_{2}$      & $T_{1}$      & $T_{3}$      & $T_{4}$      & 73.08  & 72.46  \\
    $T_{2}$      & $T_{1}$      & $T_{4}$      & $T_{3}$      & 73.10  & 72.49  \\
    $T_{3}$      & $T_{4}$      & $T_{1}$      & $T_{2}$      & 72.97  & 72.39  \\
    $T_{4}$      & $T_{3}$      & $T_{2}$      & $T_{1}$      & 73.13  & 72.49  \\
    $T_{4}$      & $T_{4}$      & $T_{4}$      & $T_{4}$      & 72.92  & 72.31  \\
    $T$  & $T$  & $T$  & $T$  & 73.02 & 72.28 \\
    \bottomrule
    \end{tabular}%
    }
  \label{tab:abseq}%
\end{table}%
According to the experimental results, we observe that incorporating contextual knowledge in the order of $[T_{1}, T_{2}, T_{3}, T_{4}]$ yields the best performance for sticker emotion understanding. Interestingly, this order also aligns with the way humans typically interpret emotional content in stickers—first considering whether the intention is sarcastic, then perceiving the overall style, followed by attention to characters and posture details. Moreover, we find that injecting only the character details ($T_{4}$) at each stage results in the lowest classification performance. This highlights the importance of our proposed multi-view description, which provides richer and more diverse contextual information, thereby enabling the model to better comprehend the emotional content embedded in the stickers. Additionally, incorporating all views at every stage also leads to performance degradation, which strongly supports the effectiveness of our proposed hierarchical fusion strategy. By guiding the model to focus on different information at different stages, this approach mitigates conflicts among multiple views and facilitates more effective cross-modal understanding.

\begin{figure}[!t]
    \centering
    \subfloat[MGHFT]{
    		\includegraphics[scale=0.225]{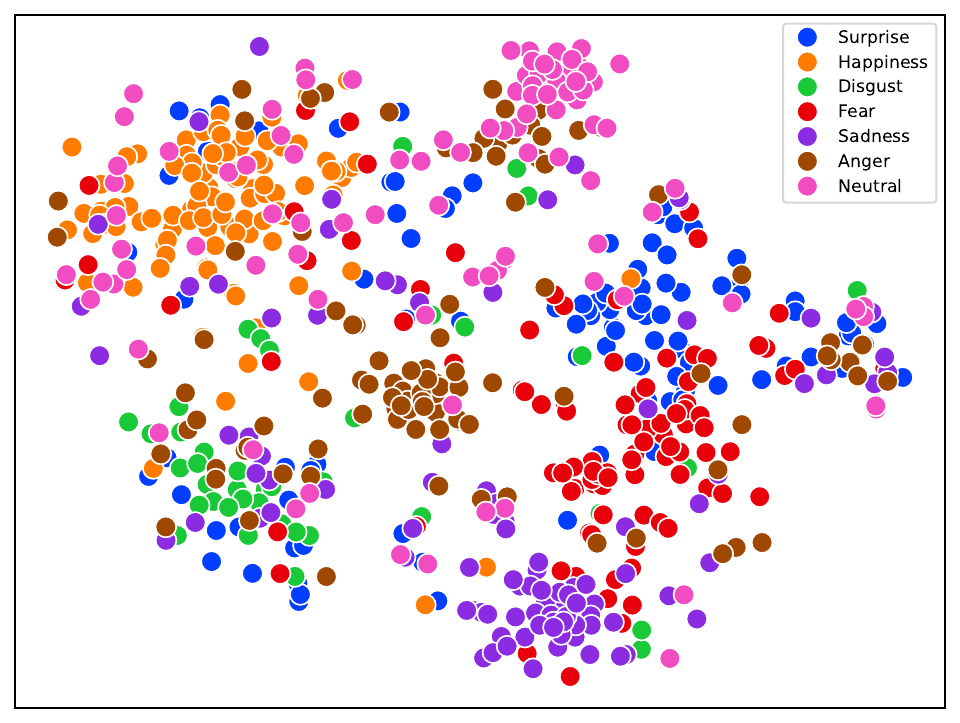}\label{fig:tsne_tgfa}}
    \subfloat[CLIP]{
    		\includegraphics[scale=0.225]{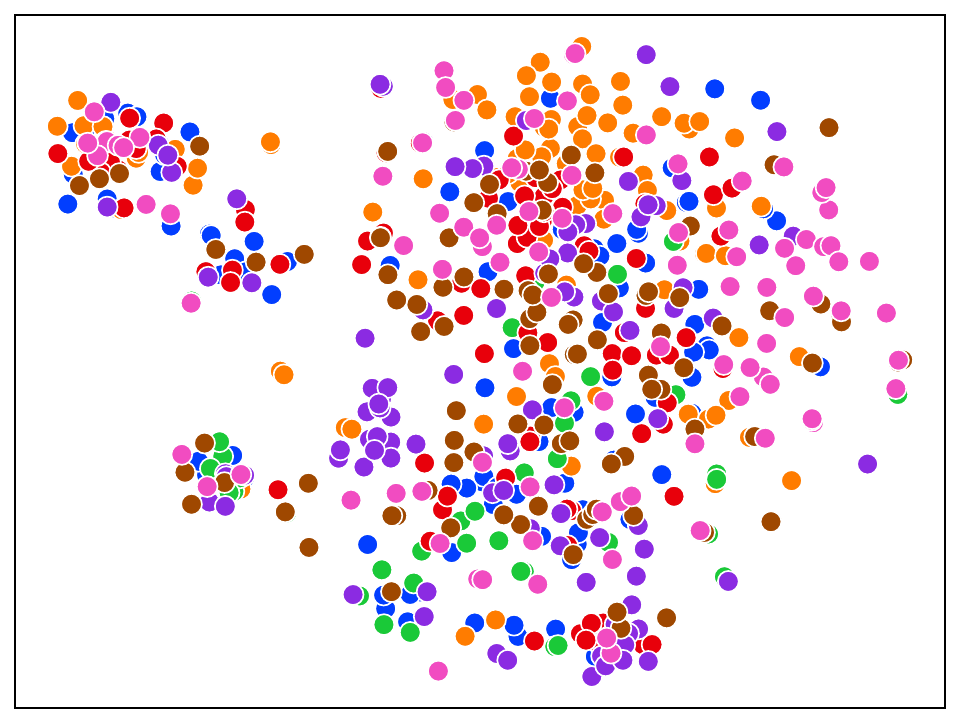}\label{fig:tsne_clip}}
    \\
    \subfloat[BLIP]{
    		\includegraphics[scale=0.225]{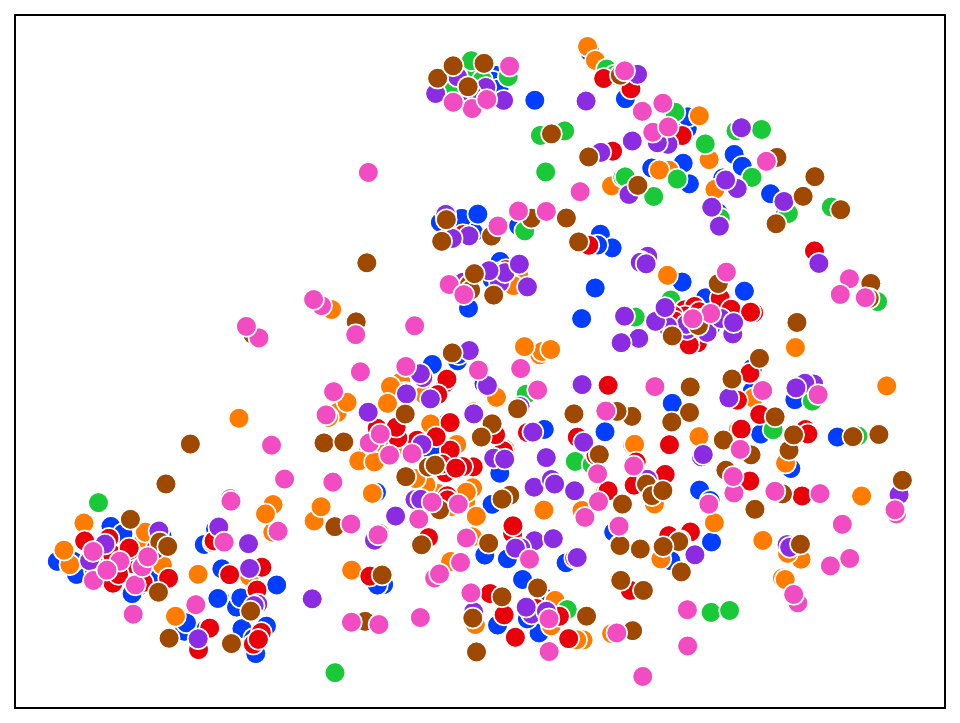}\label{fig:tsne_blip}}
    \subfloat[LLaVA]{
    		\includegraphics[scale=0.225]{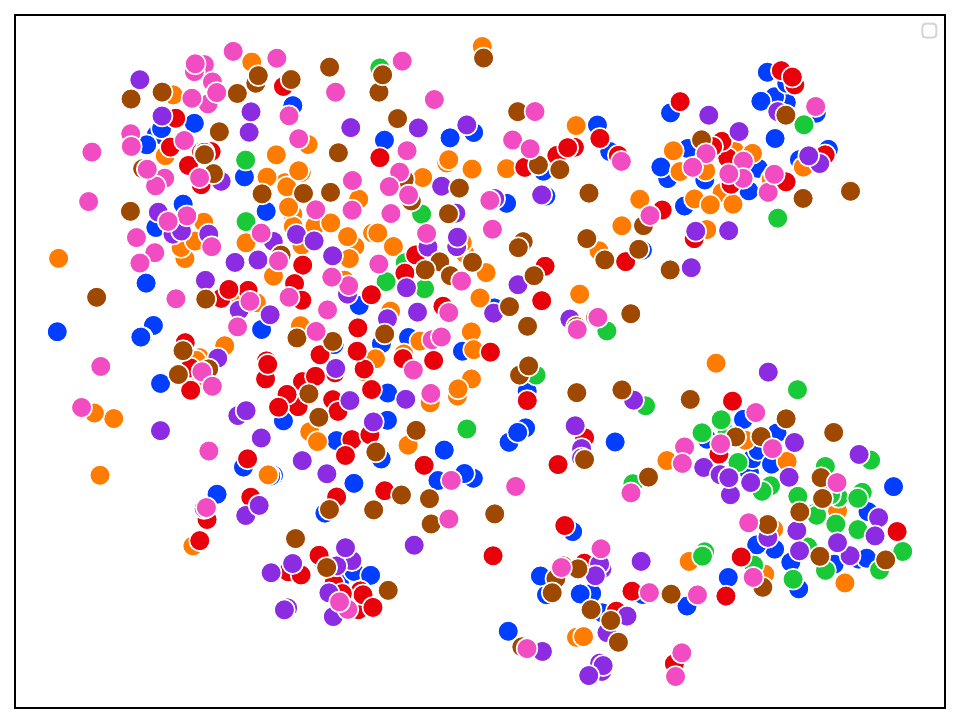}\label{fig:tsne_llava}}
    \vspace{-4.5mm}
    \caption{t-SNE visualization of 642 sticker features extracted with different methods. From the test set, we randomly sampled 100 examples per category. However, since the 'Disgust' category contains only 42 examples in the test set, all 42 available samples were used for this category. The examples are from SER30K.}
    \label{fig:tsne}
    \vspace{-10pt}
\end{figure}

\subsection{t-SNE Visualization Analysis}


We hypothesize that the performance improvement of the MGHFT model primarily stems from its more efficient sticker representation method during the learning process. To validate this assumption, we conducted a visual analysis of the sticker representations using t-distributed Stochastic Neighbor Embedding (t-SNE)~\cite{van2008visualizing}, as illustrated in Figure~\ref{fig:tsne}. The visualization results reveal that the sticker features extracted by the CLIP, BLIP and LLaVA models exhibit relatively scattered distributions with limited inter-class separability. In contrast, although the sticker features learned by the MGHFT model also maintain a degree of dispersion in the semantic space, images belonging to the same emotional category form more compact clusters. This observation suggests that, compared to conventional pre-trained vision models, our approach achieves a more distinct separation of sticker features across different emotional categories in the feature space, thereby demonstrating its superior capability in learning effective sticker representations.

\begin{figure} 
    \centering
    \includegraphics[width=0.75\linewidth]{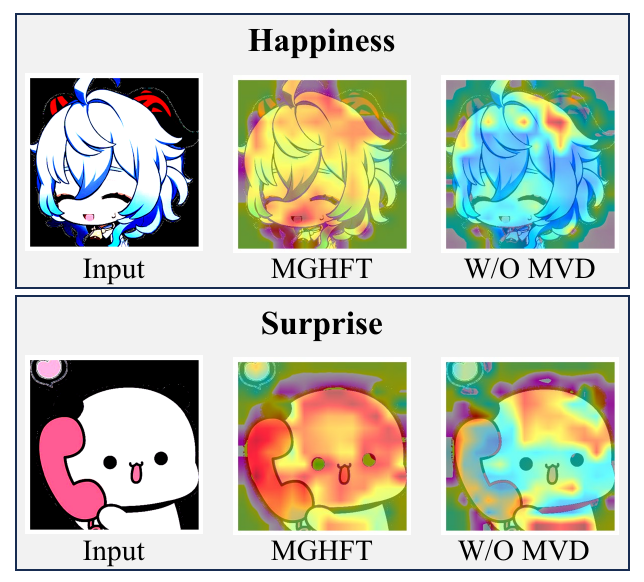}
    \vspace{-4.5mm}
    \caption{Visualization of the region of interest of the model based on last layer attention. The examples are from SER30K.}
    \label{fig:visattn}
    \vspace{-10pt}
\end{figure}

\subsection{Attention Visualization Analysis}
To better illustrate the effectiveness of our proposed MGHFT model, we visualize the model’s regions of interest using attention heatmaps and compare them with those generated by models without multi-view descriptions (W/O MVD). Specifically, we extract attention weights from each token in the final layer of the visual backbone and overlay them onto the original stickers, as shown in Figure~\ref{fig:visattn}. The visualizations clearly show that MGHFT successfully highlights the most informative regions of the stickers, such as facial expressions, eye direction, and other crucial emotional cues. In contrast, models without multi-view descriptions tend to focus on irrelevant regions. Notably, MGHFT assigns lower attention weights to semantically unimportant background areas, reflecting its ability to filter out non-informative content and concentrate on emotionally salient features. These visual comparisons provide strong evidence for the effectiveness of our multi-view textual guidance in directing the visual model’s focus toward more critical emotional features.

\section{Conclusion}



Inspired by the way humans understand the emotion of stickers, this paper proposes a novel Multi-granularity Hierarchical Fusion Transformer to overcome the limitation of existing pre-trained visual models for sticker emotion recognition. Our approach first leverages a MLLM to generate multi-view textual descriptions of stickers, providing rich semantic background knowledge to improve the model’s understanding of stickers' emotions. Built upon the PVT architecture, we innovatively integrate contrastive learning with attention mechanisms to achieve multi-granularity fusion of textual features. Furthermore, we design a text-guided multimodal attention fusion mechanism that effectively integrates visual and textual features, significantly enhancing the representational power and emotion classification accuracy of sticker data. Extensive experimental results demonstrate that each component of the proposed framework contributes meaningfully to the overall performance. We hope this work can serve as a valuable reference and inspiration for future research in sticker emotion analysis.



\bibliographystyle{ACM-Reference-Format}
\bibliography{sample-base}


\end{document}